\documentclass[runningheads]{llncs}
\usepackage[T1]{fontenc}
\usepackage{graphicx}
\usepackage{epsfig}
\usepackage{mathptmx}
\usepackage{times}
\usepackage{amsmath}
\usepackage{amssymb}
\usepackage{multirow}
\usepackage{array}
\usepackage{tabularx}
\usepackage{cite}
\usepackage{footnote}
\usepackage[hidelinks]{hyperref}
\usepackage{booktabs}
\usepackage{subcaption}
\usepackage{blindtext}
\makesavenoteenv{tabular} 

\usepackage{float}

\usepackage[x11names]{xcolor}

\begin{document}
\title{Detecting Domain Shifts in Myoelectric Activations: Challenges and Opportunities in Stream Learning}
\titlerunning{Detecting Domain Shifts in Myoelectric Activations}
%

\author{Yibin Sun\inst{1,2}\orcidID{0000-0002-8325-1889} 
\and Nick Lim\inst{2}\orcidID{0000-0003-4690-5780}
\and Guilherme Weigert Cassales\inst{2}\orcidID{0000-0003-4029-2047}
\and Heitor Murilo Gomes\inst{1}\orcidID{0000-0002-5276-637X}
\and Bernhard Pfahringer\inst{2,3}\orcidID{0000-0002-3732-5787}
\and Albert Bifet\inst{2,4}\orcidID{0000-0002-8339-7773}
\and Anany Dwivedi\inst{2}\orcidID{0000-0003-3262-6676}
}

\authorrunning{Y. Sun et al.}

\institute{
School of Engineering and Computer Science, Victoria University of Wellington, NZ.
\and
Artificial Intelligence Institute, University of Waikato, NZ. 
\and
Department of Computer Science, University of Waikato, NZ.
\and 
Big Data at Data, Intelligence and Graphs (DIG) LTCI, Télécom Paris, IP Paris, France.
\email{spencer.sun@vuw.ac.nz  anany.dwivedi@waikato.ac.nz}\\
}
\maketitle              
\begin{abstract}
Detecting domain shifts in myoelectric activations poses a significant challenge due to the inherent non-stationarity of electromyography (EMG) signals. This paper explores the detection of domain shifts using data stream (DS) learning techniques, focusing on the DB6 dataset from the Ninapro database. We define domains as distinct time-series segments based on different subjects and recording sessions, applying Kernel Principal Component Analysis (KPCA) with a cosine kernel to pre-process and highlight these shifts. By evaluating multiple drift detection methods such as CUSUM, Page-Hinckley, and ADWIN, we reveal the limitations of current techniques in achieving high performance for real-time domain shift detection in EMG signals. Our results underscore the potential of streaming-based approaches for maintaining stable EMG decoding models, while highlighting areas for further research to enhance robustness and accuracy in real-world scenarios.

\end{abstract}

\section{Introduction}\label{sec:intro}

Electromyography (EMG) is widely used for capturing neuromuscular activity by measuring the electrical potentials generated during muscle contractions \cite{de2006electromyography}. 
Surface EMG (sEMG), the non-invasive variant, is particularly advantageous due to its ease of application and high temporal resolution \cite{dwivedi2020emg, dwivedi2018emg, al2011review}. 
sEMG is commonly employed in prosthetic control, human-computer interaction, and rehabilitation systems, among other applications \cite{saponas2009enabling, artemiadis2010emg}. 
However, sEMG signals exhibit significant non-stationarity, and are influenced by factors such as electrode displacement, muscle fatigue, skin impedance variations, and external noise \cite{jiang2020novel}. 
These temporal variations introduce domain shifts in EMG data, progressively degrading the performance of intention decoding models and compromising the long-term reliability of EMG-based systems. As performance deteriorates, users experience increased frustration, ultimately leading to device abandonment and rejection~\cite{eddy2023framework}.
One way to address this challenge is through recalibration, which implies adapting the decoding models using data collected under domain shifted conditions. 
A critical first step toward this goal is the real-time detection of domain shifts, an area that remains relatively under-explored. 
Since EMG signals are streamed in real-time applications, managing domain shifts requires adaptive methods that can continuously learn and update from incoming data without interrupting user experience.

Researchers have proposed various interpretations of the term ``\textbf{\textit{domain}}’’ in the context of myoelectric signals. For example, \cite{ref_subject_domain} considers data collected from different subjects as distinct domains. Building on this, \cite{ref_time_domain} introduces a finer-grained view by separating domains based on the time of data collection—such as morning and afternoon sessions in DB6. Furthermore, cross-day experiments discussed in~\cite{ref_slot_domain} show that signal patterns can also vary significantly across days. In this work, we adopt a comprehensive definition: each unique combination of subject, day, and time slot constitutes a separate domain.

Over the past decades, data stream (DS) learning has emerged as a prominent paradigm for handling continuous, evolving, and potentially unbounded data~\cite{ref_ds}. DS algorithms enable learning from data in real-time without requiring repeated access to past observations. This has made DS techniques applicable across various domains, including financial transactions, online monitoring, and sensor-based systems—such as EMG signal analysis.

Given the non-stationary nature of EMG signals, DS learning provides a natural framework for managing their variability. In particular, it supports dynamic model adaptation through \textbf{drift detection} techniques~\cite{ref_moa}, helping to mitigate the effects of domain shifts and enhance long-term stability~\cite{lu2018learning}.

In this study, we investigate domain shift detection in EMG data using the DB6 dataset from the widely used Ninapro database~\cite{ref_db6}. Each time-series segment corresponding to a specific subject and time slot is treated as a distinct \textbf{\textit{domain}}. We demonstrate the presence of inter-domain variation using preprocessing techniques such as RMS normalization and Kernel PCA. Then, we evaluate the formulated data in a fully incremental setting, examining the effectiveness of existing drift detection methods in capturing domain shifts in real time.

Streaming solutions offer a key advantage over traditional machine learning approaches when working with time varying data due to their incremental nature. Traditional machine learning models remain fixed, once trained, and lack the ability to adapt to evolving conditions. As a result, subtle differences caused by muscle fatigue, sensor misplacement, or a change in the user often lead to errors. In contrast, streaming methods continuously update the model, allowing them to identify these changes as they occur and adjust accordingly, ensuring sustained performance over time.

This paper contributes to the detection of domain shifts in myoelectric (EMG) data using data stream (DS) learning techniques. We define domains as distinct time-series segments based on different subjects, recording sessions, and times of the day. Kernel Principal Component Analysis (KPCA) with a cosine kernel is applied for feature extraction and dimensionality reduction. We comprehensively evaluate multiple drift detection methods, revealing limitations in their performance. Finally, we discuss key challenges and propose future research directions, such as developing incremental non-linear decomposition models and integrating complementary sensor data to enhance robustness and accuracy.

The rest of the paper is organized as follows: Section~\ref{sec:Relatedwork} presents the related work, followed by a exhibitions of the preliminary analysis in Section~\ref{sec:Method}. The experiments and results are covered in Section~\ref{sec:Results}. Finally, Section~\ref{sec:Conclusion} concludes this work and proposes future research directions.

\section{Related Work}\label{sec:Relatedwork}

Electromyographic (EMG) pattern recognition systems have demonstrated high classification accuracies in controlled laboratory settings. 
However, their robustness over time remains a significant challenge. 
Studies have shown that the performance of these systems can degrade within hours after initial classifier training, primarily due to variations in EMG signals caused by factors such as electrode displacement, muscle fatigue, and changes in muscle contraction effort~\cite{sensinger2009adaptive}.
To mitigate these issues, researchers have explored various strategies. One approach involves identifying EMG features that are less sensitive to such disturbances while maintaining high class separability. For instance, time-domain features like autoregression coefficients and cepstrum coefficients have been found to offer robust classification performance under varying conditions~\cite{tkach2010study}.

Another strategy focuses on adaptive learning methods to counteract performance degradation. 
A Bayesian approach for adaptive EMG pattern classification using semi-supervised sequential learning has been proposed, which updates the model to accommodate alterations in signal characteristics over time~\cite{yoneda2023bayesian}.
Additionally, domain adaptation techniques have been investigated to address the non-stationarity of EMG signals. For example, EMGSense is a framework that tackles performance degradation caused by time-varying biological and environmental factors through self-supervised domain adaptation~\cite{duan2023emgsense}. 
Despite these advancements, achieving long-term stability in EMG-based systems remains an ongoing area of research. Continuous efforts are directed towards developing methods that can dynamically adapt to signal variations, ensuring reliable performance in real-world applications. 

In this context, Principal Component Analysis (PCA) can be a powerful tool to identify distribution shifts. In statistics, PCA is often used to reveal the main axes of variation in a dataset, and comparing how two datasets project onto these axes can highlight differences in their underlying distribution~\cite{jollifePCA, kernelPCA}. Essentially, if one dataset occupies a distinct region in the principal component space, or exhibits different variances along the key components, this indicates a distribution shift. In our approach we use Kernel PCA with a cosine kernel,
$k(x,y) = \dfrac{\mathbf{x}\cdot \mathbf{y}}{\|\mathbf{x}\|\| \mathbf{y}\|}$,
such that the emphasis is on the angular relationships between the instances in the time series, which can be a powerful approach for comparing signals~\cite{hoffmanKPCA,SmolakKPCA}. Such approaches have shown some success detecting out-of-distribution and distribution shifts in machine learning datasets in works by~\cite{guanPCAOOD,fangPCAOOD}. 

Complicating this further however is the fact, that these variations often manifest as gradual or sudden changes in the data distribution, known as concept drifts in stream learning. As the changes happen progressively without the researcher's awareness, detecting and addressing these concept drifts (CD) becomes essential for maintaining model performance over time. 

\subsection{Drift Detectors}

One of the fundamental approaches to stream learning is the detection of concept drifts. The literature has proposed a sizable number of drift detection methods~\cite{ref_cd,lu2018learning}. In this work, we highlight the most widely adopted and well-established methods.

The \textbf{CUSUM} (Cumulative Sum)~\cite{ref_cusum} algorithm detects shifts in the mean of a data stream (DS) by continuously accumulating the deviations of observed values from an estimated baseline or reference value. When the cumulative sum exceeds a predefined threshold, it signals a potential concept drift, indicating that the underlying distribution of the data may have changed.

Similar in intent but with a different smoothing mechanism, the \textbf{GMA} (Geometric Moving Average)~\cite{ref_ewma} method tracks changes by computing an exponentially weighted moving average of the deviations. This approach gives more weight to recent observations, making it particularly effective for identifying sudden but short-lived shifts in the data.

Another well-known technique is the \textbf{Page-Hinckley} (PH) test~\cite{ref_phtest}, which emphasizes detecting sustained, gradual deviations by tracking the cumulative difference between observed values and their running average. It signals drift when this cumulative deviation consistently exceeds a tolerance level, making it particularly effective for identifying slower shifts in data streams. 

The \textbf{DDM} (Drift Detection Method)~\cite{ref_ddm} offers a different perspective by monitoring the online error rate of a learning model rather than tracking statistical properties of the data stream directly. DDM assumes that the error rate will decrease or remain stable as the model learns. A significant increase in the error rate, beyond statistically defined thresholds, indicates potential drift and prompts model adaptation. 

The \textbf{ADWIN} (Adaptive Windowing) algorithm~\cite{ref_adwin} adopts a dynamic windowing approach to concept drift detection. Instead of relying on fixed thresholds, ADWIN maintains a variable-length sliding window of recent observations and continuously compares statistics between sub-windows. When a statistically significant difference is detected between two segments of the window, ADWIN concludes that a change has occurred, automatically shrinking the window to retain only the most recent, relevant data. This adaptability makes it well-suited for non-stationary environments with evolving distributions.
 
The \textbf{HDDM$_A$} (Average Hoeffding’s Bounds Drift Detection Method)~\cite{ref_hddm} leverages Hoeffding’s inequality to statistically monitor the mean of a data stream over time. By maintaining an average error rate and estimating the confidence interval around it, HDDM$_A$ detects concept drift when a significant deviation exceeds the calculated bound, indicating that the distribution of the stream may have changed.

Building on this, the \textbf{HDDM$_W$} (Weighted Hoeffding’s Bounds Drift Detection Method) introduces a weighting scheme that emphasizes more recent observations. This enhancement enables the detector to respond more quickly to emerging drifts while maintaining robustness against transient fluctuations, thereby improving both sensitivity and precision in dynamic environments.

The \textbf{SEED} (Stream-based Efficient and Effective Drift) detector~\cite{ref_seed} is a lightweight and high-speed drift detection method designed specifically for streaming environments. It operates by dividing incoming data into fixed-size blocks and applying statistical hypothesis testing to compare the distributions of recent and past blocks. This block-based strategy enables SEED to efficiently detect both abrupt and gradual changes with low computational and memory overhead. By avoiding the need to maintain long sliding windows or continuously update complex statistics, SEED offers a scalable and robust solution for real-time drift detection, while also incorporating safeguards to reduce false positives.

More recently, the \textbf{Adaptive Bernstein Change Detector (ABCD)}~\cite{ref_abcd} has been proposed to address some of the limitations of earlier detectors by incorporating variance-sensitive statistical bounds. ABCD uses the Bernstein inequality to dynamically estimate confidence intervals around the empirical mean, enabling robust and theoretically grounded drift detection. By adapting to both the observed variance and sample size, ABCD can effectively detect a wide range of drift types while minimizing false alarms, making it suitable for complex, evolving data streams.

All these detectors are available in \textbf{\href{https://capymoa.org}{CapyMOA}}~\cite{ref_capymoa} and MOA~\cite{ref_moa}, both of which are open-source platforms for data stream learning.

\section{Preliminary Analysis}\label{sec:Method}

\subsection{Preprocessing}

In our preprocessing pipeline, we begin by eliminating columns that contain only zero values throughout the recording. This step is particularly relevant for the EMG data in the NinaPro DB6 dataset, where column 9 and column 10 consistently contain zero readings and therefore provide no useful signal for downstream analysis.

Following column elimination, we apply a sliding window approach to extract the root mean squared (RMS) value as a feature representation. RMS is a widely used measure in EMG signal processing, as it captures the signal’s energy and correlates well with muscle activation levels. We use a window size of 200 ms with a stride of 20 ms, striking a balance between temporal resolution and noise reduction. This transformation effectively smooths the high-frequency noise inherent in EMG recordings while preserving local variations essential for distinguishing different gesture patterns.

Together, these preprocessing steps serve to clean and compact the raw signal data, providing a more informative and tractable input for subsequent modeling stages.

\subsection{PCA Analysis}
Following this, we employ Kernel Principal Component Analysis (Kernel PCA) with a Cosine kernel to project the data into a lower-dimensional space with three principal components, capturing the most relevant structural patterns. While we also experimented with alternative kernels, such as the Radial Basis Function (RBF), our results indicate that the overall trends remain consistent across different kernels.

\begin{figure}[!htbp]
    \centering
    \begin{subfigure}[b]{0.45\textwidth}
        \centering
        \includegraphics[width=\textwidth]{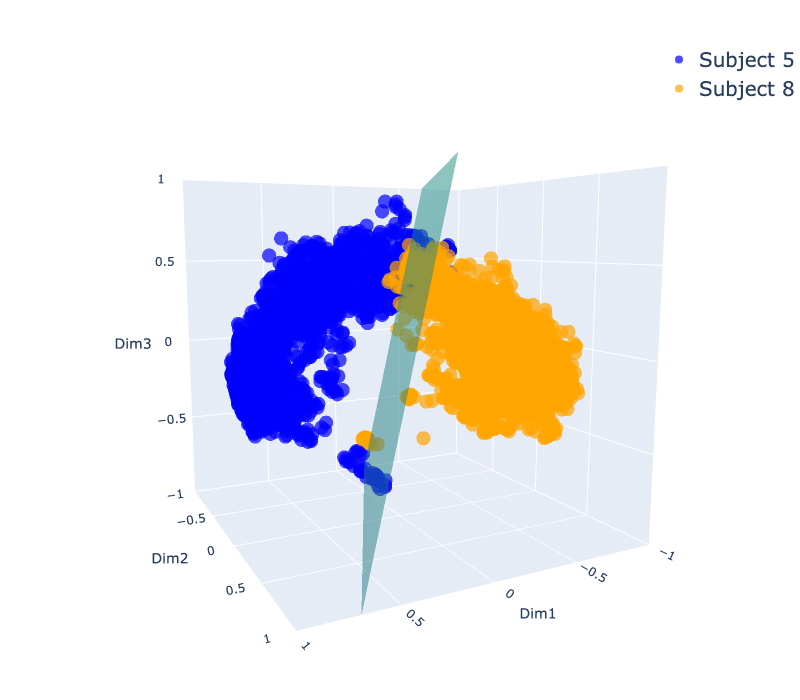}
        \caption{Grasp 1}
    \end{subfigure}
    \hfill
    \begin{subfigure}[b]{0.45\textwidth}
        \centering
        \includegraphics[width=\textwidth]{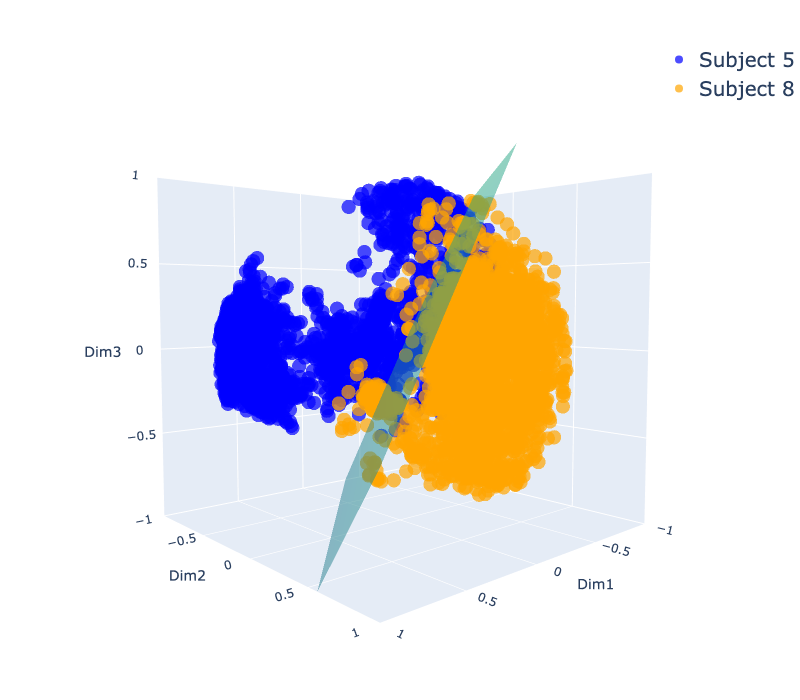}
        \caption{Grasp 3}
    \end{subfigure}
    
    \begin{subfigure}[b]{0.45\textwidth}
        \centering
        \includegraphics[width=\textwidth]{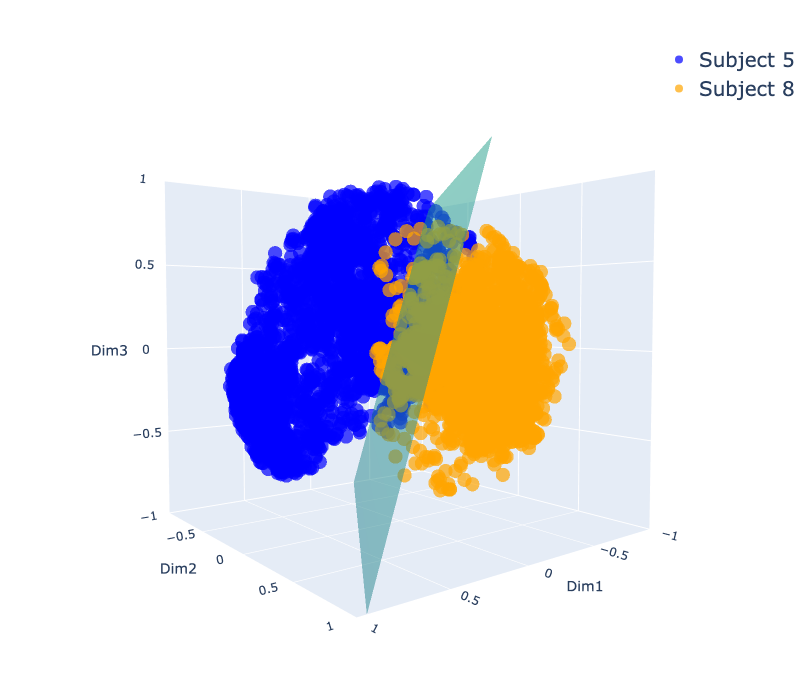}
        \caption{Grasp 4}
    \end{subfigure}
    \hfill
    \begin{subfigure}[b]{0.45\textwidth}
        \centering
        \includegraphics[width=\textwidth]{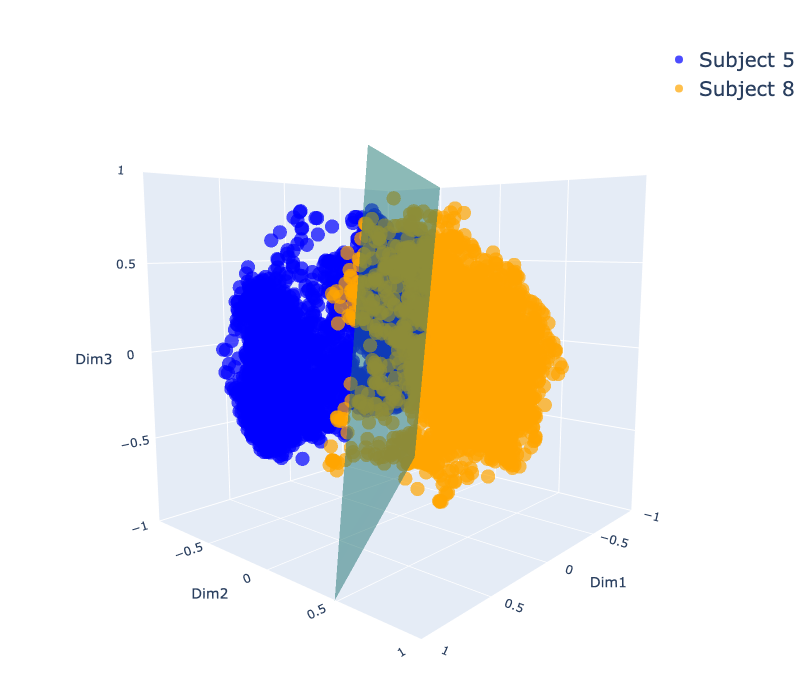}
        \caption{Grasp 6}
    \end{subfigure}

    \begin{subfigure}[b]{0.45\textwidth}
        \centering
        \includegraphics[width=\textwidth]{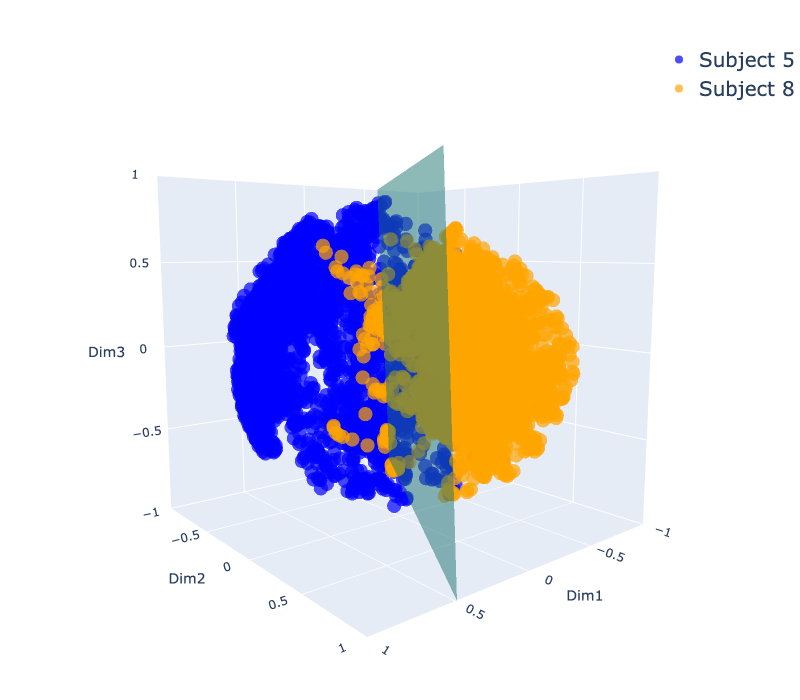}
        \caption{Grasp 9}
    \end{subfigure}
    \hfill
    \begin{subfigure}[b]{0.45\textwidth}
        \centering
        \includegraphics[width=\textwidth]{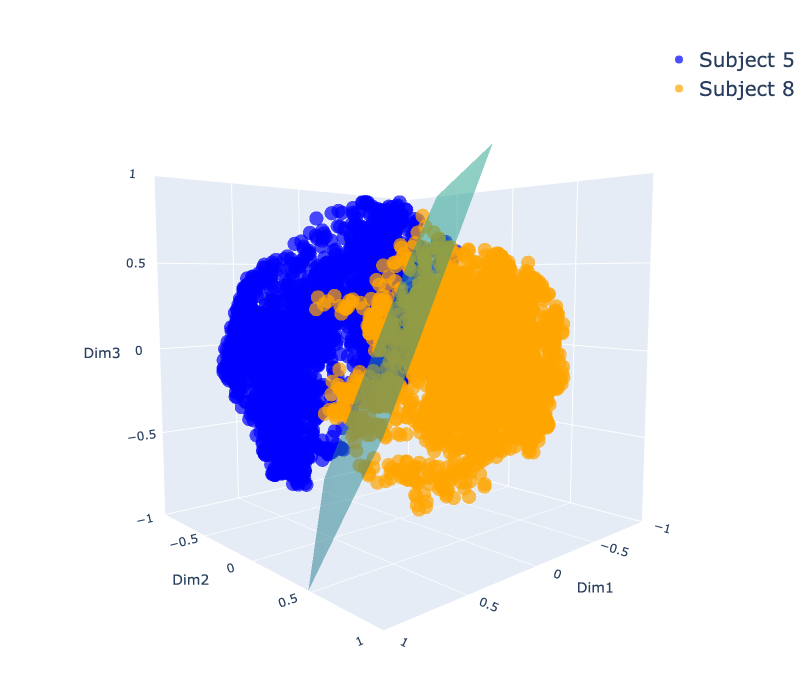}
        \caption{Grasp 10}
    \end{subfigure}
    
    \begin{subfigure}[b]{0.45\textwidth}
        \centering
        \includegraphics[width=\textwidth]{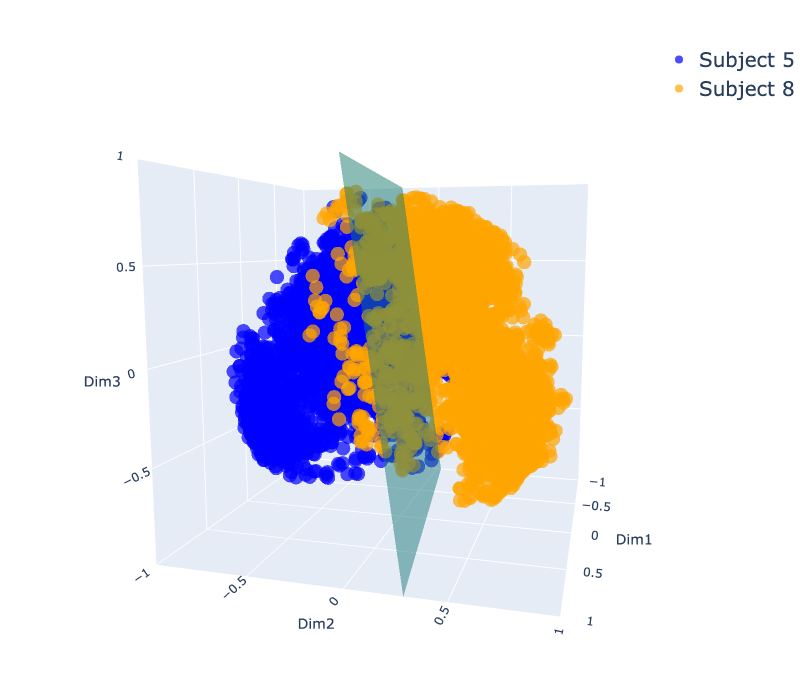}
        \caption{Grasp 11}
    \end{subfigure}

    \caption{Feature Extracted Visualizations from \textbf{Subject 5 and 8} at Period 5}
    \label{fig:different_subjects}
\end{figure}

\begin{figure}[!htbp]
    \centering
    \begin{subfigure}[b]{0.45\textwidth}
        \centering
        \includegraphics[width=\textwidth]{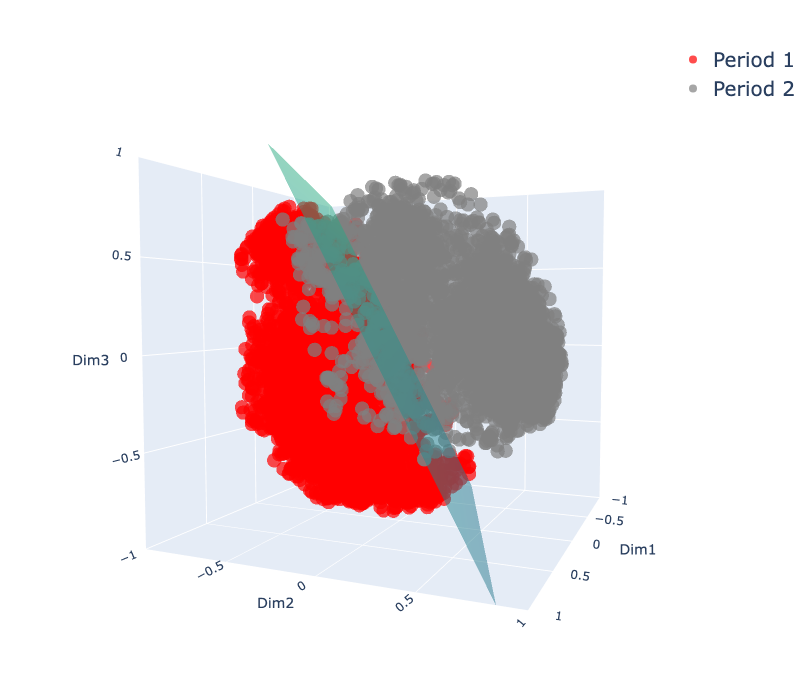}
        \caption{Grasp 1}
    \end{subfigure}
    \hfill
    \begin{subfigure}[b]{0.45\textwidth}
        \centering
        \includegraphics[width=\textwidth]{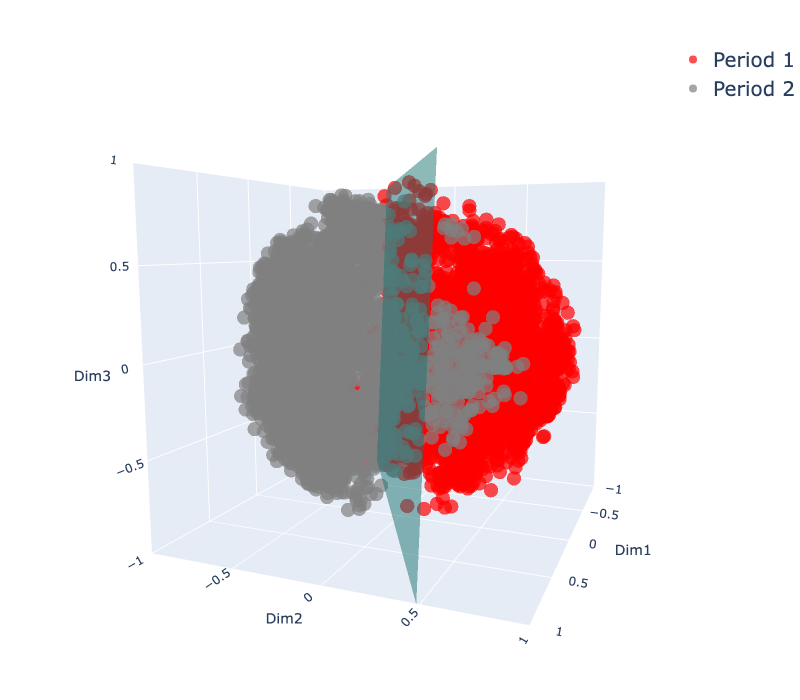}
        \caption{Grasp 3}
    \end{subfigure}

    \begin{subfigure}[b]{0.45\textwidth}
        \centering
        \includegraphics[width=\textwidth]{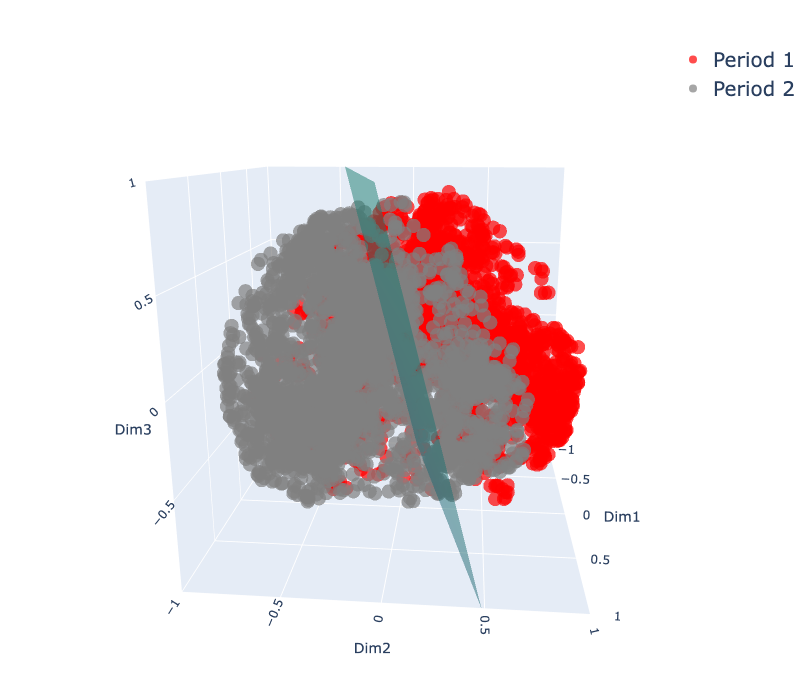}
        \caption{Grasp 4}
    \end{subfigure}
    \hfill
    \begin{subfigure}[b]{0.45\textwidth}
        \centering
        \includegraphics[width=\textwidth]{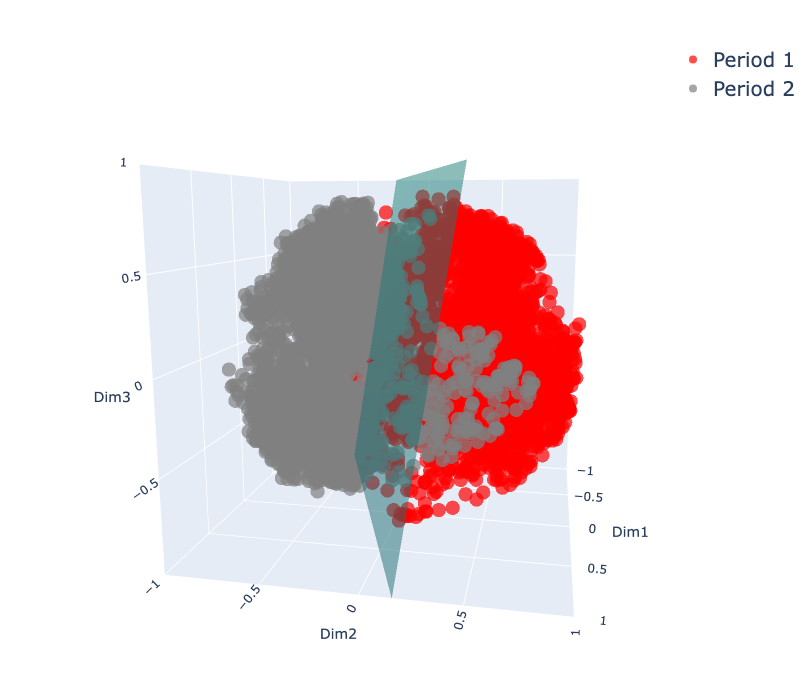}
        \caption{Grasp 6}
    \end{subfigure}

    \begin{subfigure}[b]{0.45\textwidth}
        \centering
        \includegraphics[width=\textwidth]{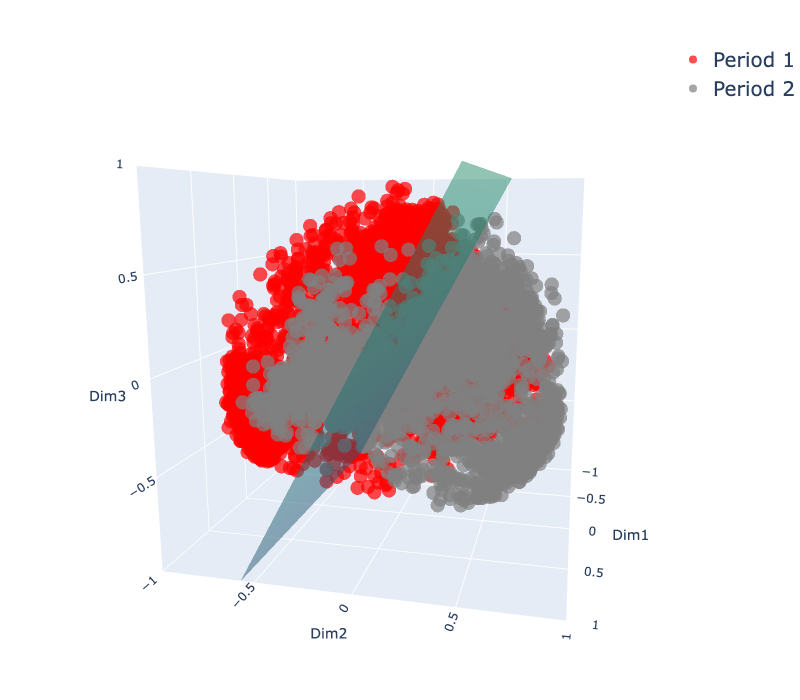}
        \caption{Grasp 9}
    \end{subfigure}
    \hfill
    \begin{subfigure}[b]{0.45\textwidth}
        \centering
        \includegraphics[width=\textwidth]{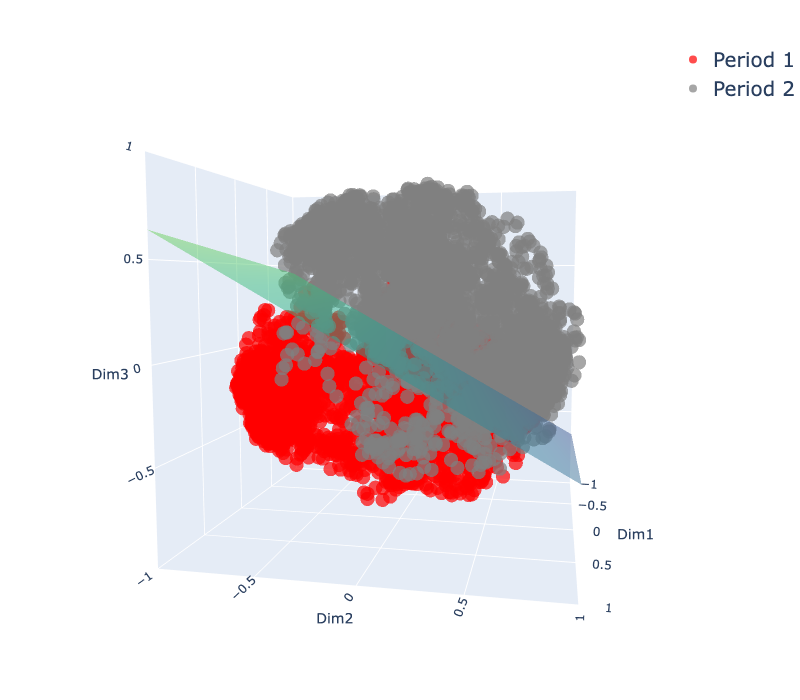}
        \caption{Grasp 10}
    \end{subfigure}
    
    \begin{subfigure}[b]{0.45\textwidth}
        \centering
        \includegraphics[width=\textwidth]{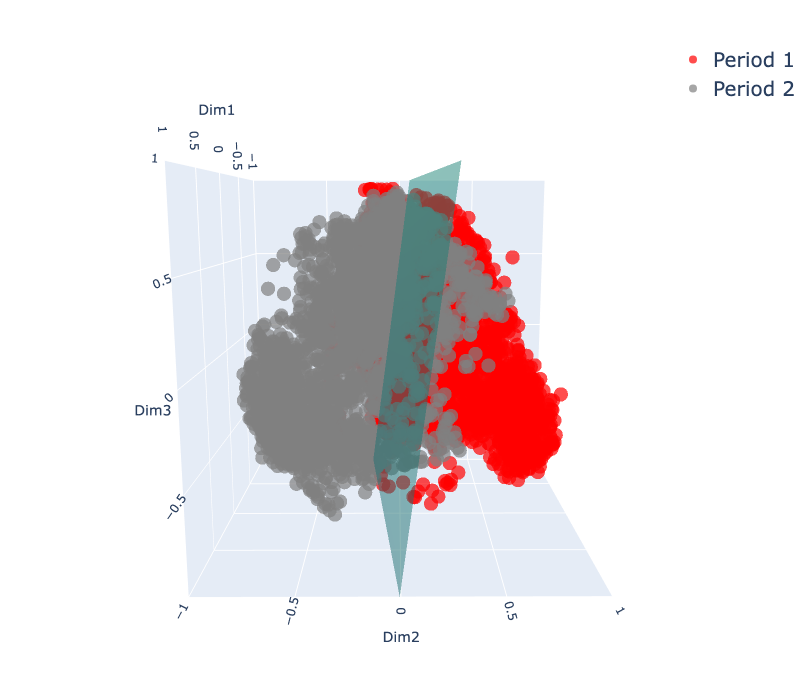}
        \caption{Grasp 11}
    \end{subfigure}

    \caption{Feature Extracted Visualizations from Subject 1 at \textbf{Period 1 and 2}}
    \label{fig:different_times}
\end{figure}

Fig.~\ref{fig:different_subjects} and Fig.~\ref{fig:different_times} illustrate the preprocessing outcomes. The hyperplanes in these figures represent logistic regression classification boundaries.

Since Ninapro's DB6 includes data from 5 days with 2 time slots per day per subject, we simplify the time notation as follows: T1 at D1 becomes Period 1 (P1), T2 at D1 becomes Period 2 (P2), T1 at D2 becomes Period 3 (P3), and so on.

In Fig.~\ref{fig:different_subjects}, we show two \textbf{different subjects} (Subject~5 and Subject~8) for P5.
Meanwhile, Fig.~\ref{fig:different_times} depicts the same subject performing identical gestures at \textbf{different time periods} on the same day.

From both the figures, it can be seen that the data for all grasp categories is distinctly segmented into two well-separated groups, as indicated by the classification hyperplanes.
The fact that the samples lie on two distinct hemispheres is interesting for several reasons. 
First, it suggests that there is a substantial difference between the signals, possibly reflecting genuine changes in the physiological signals. 
In other words, even if the raw EMG data looked visually simple, the cosine kernel captures the angular difference that shows clean separation in the kernel-induced feature space. 

Second, the hemispheric split may indicate a distribution shift that could pose challenges for models trained solely on one group of data (morning vs evening data recordings, see Fig.~\ref{fig:different_times}). 
While not all plots in Fig.~\ref{fig:different_times} exhibit clearly distinguishable sections, the majority are properly segmented. However, Grasp 4 and Grasp 9 display more overlapping patterns, making it more challenging to detect them as instances of domain shift or concept drift.
Any model ignoring this shift may suffer performance degradation when confronted with the other set (e.g. model trained in the morning will have deteriorated performance in the evening). As a result, researchers should consider strategies like recalibration, domain adaptation or a separate model for each sub group to maintain accuracy.

Finally, from a feature-learning perspective, seeing a distinct separation underscores the power of non-linear dimensionality reduction. This paper reflects a preliminary work using this approach, it is possible that there are additional implications that we have not considered and exploited that would be useful for domain adaptation in EMG signal.

Based on the analysis on the illustrations in Fig.~\ref{fig:different_subjects} and~\ref{fig:different_times}, significant differences are observed across different time periods and among different subjects. In this study, these characteristics will be collectively defined as domain semantics. Accordingly, we concatenate the data from different time periods and subjects, forming a dataset that serves as a stream for the experiments in the subsequent sections.

\subsection{Reference-Based Distribution Drift Analysis}
\label{sec:drift_analysis}

While Principal Component Analysis (PCA) provides useful insights into the structure of the data, it relies on the assumption that the dataset is stationary—that all samples come from a fixed, unchanging distribution. In practice, especially with sEMG signals, this assumption rarely holds. The underlying data distribution often changes over time due to factors such as electrode shifts, perspiration, and muscle fatigue, making the analysis more complex and less reliable if these shifts are ignored.

To capture and quantify such temporal dynamics, we apply a Reference-Based Distribution Drift Analysis. In this approach, an initial stable segment of the signal is modeled as a reference distribution. The remainder of the time series is then analyzed through sliding windows, each compared to the reference. This method enables the detection and characterization of gradual or abrupt distributional shifts that would be invisible to static, batch-oriented methods such as PCA.

\subsubsection{Reference Window Modeling}
A reference window is first selected from an initial stable segment of the data. The signals within this window are assumed to follow a multivariate Gaussian distribution, characterized by a mean vector $\boldsymbol{\mu}_0 \in \mathbb{R}^d$ and a covariance matrix $\boldsymbol{\Sigma}_0 \in \mathbb{R}^{d \times d}$, where $d$ is the number of signal channels. The parameters are estimated as:

\begin{equation}
\boldsymbol{\mu}_0 = \frac{1}{N} \sum_{i=1}^{N} \mathbf{x}_i, \quad
\boldsymbol{\Sigma}_0 = \frac{1}{N-1} \sum_{i=1}^{N} (\mathbf{x}_i - \boldsymbol{\mu}_0)(\mathbf{x}_i - \boldsymbol{\mu}_0)^\top,
\end{equation}
where $\mathbf{x}_i \in \mathbb{R}^d$ are the individual samples from the reference window.

\subsubsection{Sliding Window Distribution Comparison}
Subsequently, a sliding window is moved across the remainder of the data with a specified step size. For each window, a local Gaussian model $\mathcal{N}(\boldsymbol{\mu}_1, \boldsymbol{\Sigma}_1)$ is fitted.

To quantify the difference between the local distribution and the reference, we compute the Kullback-Leibler (KL) divergence between the two multivariate Gaussian distributions. The closed-form KL divergence between two Gaussian distributions is given by~\cite{cover1999elements}:

\begin{equation}
D_{\text{KL}}(\mathcal{N}_0 \,||\, \mathcal{N}_1) = \frac{1}{2} \left( \mathrm{tr}(\boldsymbol{\Sigma}_1^{-1} \boldsymbol{\Sigma}_0) + (\boldsymbol{\mu}_1 - \boldsymbol{\mu}_0)^\top \boldsymbol{\Sigma}_1^{-1} (\boldsymbol{\mu}_1 - \boldsymbol{\mu}_0) - d + \ln\left(\frac{\det \boldsymbol{\Sigma}_1}{\det \boldsymbol{\Sigma}_0}\right) \right),
\end{equation}
where $\det(\cdot)$ denotes the matrix determinant.

\begin{figure}
    \centering
\includegraphics[width=\linewidth]{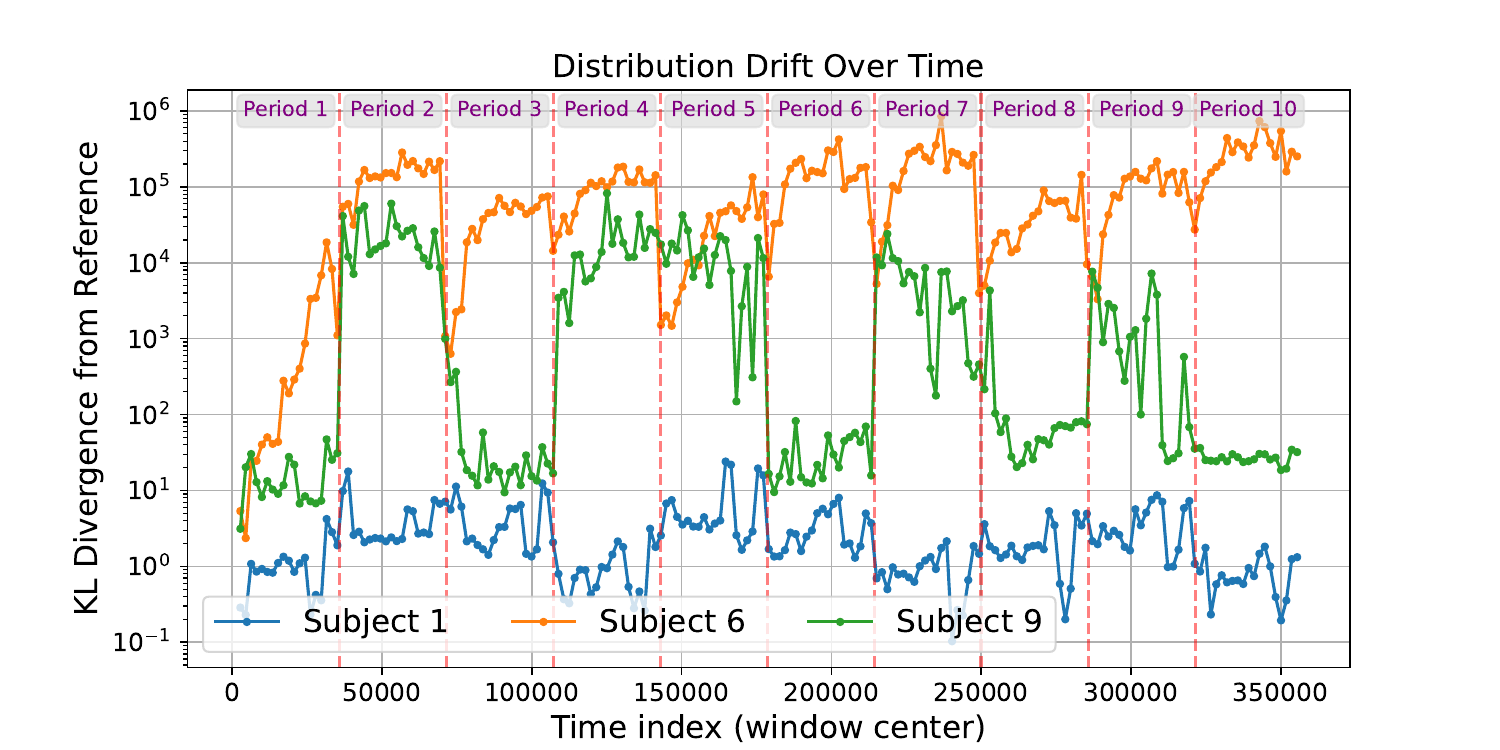}
    \caption{KL divergence between the sliding windows and the initial reference distribution over time.
The curve illustrates the degree of distributional drift relative to the baseline signal characteristics. Periods of low divergence suggest stability, while localized peaks may indicate potential changes in signal behavior.
Red dashed vertical lines mark the points of transitions among subjects and periods of the day.}
\label{fig:kl_divergence}
\end{figure}

The KL divergence plot presented in Figure~\ref{fig:kl_divergence} shows the evolution of distributional changes in the signal over different periods with respect to the initial reference window from Subject 1, 6, and 9. 
Noticeably, the y-axis is in log-scale in Figure~\ref{fig:kl_divergence}. Note that only these three subjects are exhibited here since other subjects showed more vague patterns. Both window size and step length are set at 1600, making sure that most of the data points in the window are from the same gesture\footnote{After the data extraction, each gesture is lasting for roughly 3200 data points.}.

Overall, the KL divergence seldom stays relatively low and stable during most of the recording, suggesting that the data distribution significantly deviate from the reference under normal conditions. 
Furthermore, the tendencies differ from each other largely for different subjects. In the selected subjects in Figure~\ref{fig:kl_divergence}, sudden shifts that are aligned with the period changes can be clearly observed during the transition from one period to another, indicating that the sEMG signals for different periods, even from the same performer, show significant divergence. 

On another side, obvious difference can also be seen in the KL-Divergence values between different subjects. For example, subject 1 (blue line) produced lower divergence while subject 6 (orange line) demonstrates sustainably large values. This phenomenon confirms that different people activate muscle in relatively different manners even when they are doing similar movements.

Even though not all the subjects and periods illustrate clear and distinguishable patterns over time under our preliminary analyses, we can still assume that under certain situation, the domain shifts between different subjects and periods can be detected in real-time applications.
Based on this assumption, along with the well-established drift detectors in stream learning (see Section~\ref{sec:Relatedwork}), we conducted experiments described in the following section.

\section{Drift Detection Results}\label{sec:Results}

A comprehensive set of experiments were conducted using well-established drift detection methods under multiple approaches of preprocessing. Most of the results show an extremely high False Positive Rate (FPR), staying above 90\% even when tuning the sensitivity. 
After experimenting with several configurations, the best results were achieved by applying a specific pipeline that can be implemented in a streaming manner: 
\textbf{1)} We extract RMS from the raw EMG signal using windows of 200 ms with 20 ms stride; 
\textbf{2)} From the extracted RMS signal, we create windows of 1500 data points with a stride of 500 (i.e., subsequent windows will have 1000 overlapping data points);
\textbf{3)} We fit a linear least-squares regression on the window and extract the slope; 
\textbf{4)} The slopes are used to compute the Mahalanobis distance with a rolling reference window, estimating the distance (score) between two consecutive movements; \textbf{5)} Finally, the drift detector algorithm monitors the scores.

Notably, \textbf{this research considers the tasks as unsupervised} -- i.e., no labels were provided to the algorithms -- to better align with real-world scenarios.

\begin{table}[!ht]
\centering
\caption{F1 Score and Average Detection Delay (\texttt{ADD}) in Seconds Results from Commonly Used Detectors}
\vspace{10pt}
\label{tab:result}

\begin{tabular}{l|c|rrrrrrr}
\toprule
\textsc{Detector}& \textsc{Metrics} &Grasp 1 & Grasp 3 & Grasp 4 & Grasp 6 & Grasp 9 & Grasp 10 & Grasp 11\\\midrule
\multirow{2}{*}{ADWIN}  & F1  & 0.019   & --       & 0.02    & 0.019   & 0.02    & 0.02    & 0.019   \\
                        & \texttt{ADD}  & 1.62      & --       & 8.76     & 3.40     & 3.88     & 7.40     & 8.26     \\\midrule
\multirow{2}{*}{CUSUM}  & F1  & 0.018   & 0.035   & 0.051   & --       & 0.034   & 0.035   & 0.035   \\
                        & \texttt{ADD}  & 5.66     & 6.74     & 2.32     & --       & 7.80     & 6.46     & 7.60     \\\midrule
\multirow{2}{*}{GMA}    & F1  & --       & 0.02    & 0.019   & 0.019   & 0.019   & 0.019   & 0.019   \\
                        & \texttt{ADD}  & --       & 5.42     & 6.94     & 7.72     & 7.72     & 6.32     & 8.26     \\\midrule
\multirow{2}{*}{HDDM$_A$} & F1  & \textbf{0.409}  & 0.35    & 0.376   & \textbf{0.317}   & \textbf{0.341}   & \textbf{0.347}   & 0.345   \\
                        & \texttt{ADD}  & 4.76     & 4.84     & 5.36     & 4.46     & 4.66     & 5.22     & 5.10     \\\midrule
\multirow{2}{*}{HDDM$_W$} & F1  & 0.394   & \textbf{0.352}   & \textbf{0.378}   & 0.311   & 0.34    & 0.345   & \textbf{0.349}   \\
                        & \texttt{ADD}  & 4.80     & 4.84     & 5.22     & 4.70     & 4.58     & 5.24     & 5.16     \\\midrule
\multirow{2}{*}{PH}     & F1  & --       & 0.02    & 0.02    & --       & --       & --       & --       \\
                        & \texttt{ADD}  & --       & 5.42     & 2.28     & --       & --       & --       & --       \\\midrule
\multirow{2}{*}{SEED}   & F1  & 0.038   & 0.039   & 0.038   & 0.057   & 0.019   & --       & 0.019   \\
                        & \texttt{ADD}  & 3.08     & 7.64     & 4.58     & 4.64     & 3.88     & --       & 8.26     \\\midrule
\multirow{2}{*}{ABCD}   & F1  & 0.02    & --       & --       & --       & 0.02    & --       & --       \\
                        & \texttt{ADD}  & 4.18     & --       & --       & --       & --       & --       & --       \\\bottomrule
\multicolumn{9}{l}{\footnotesize{\textsc{$^{*}$Grasp 2, 5, 7, and 8 are not included within Ninapro's DB6 database.}}}
\end{tabular}
\end{table}

Table~\ref{tab:result}~\footnote{The results for DDM~\cite{ref_ddm}, EWMA~\cite{ref_ewma}, RDDM~\cite{ref_rddm}, and STEPD~\cite{ref_stepd} were also evaluated, but none of these methods detected any changes correctly. Therefore, their results are omitted from the table for clarity.} summarizes the results from the above described experiments. Both F1 score and Average Detection Delay (\texttt{ADD}) are included for each available Grasp category across various drift detectors.

The results in Table~\ref{tab:result} highlight several key observations regarding the performance of commonly used drift detection algorithms. Overall, most detectors exhibit poor performance in terms of F1 score. The best-performing detectors, HDDM$_A$ and HDDM$_W$, demonstrate relatively higher F1 scores compared to others, with maximum values of 0.409 and 0.378, respectively. Despite being the best-performing among the tested detectors, these scores are still far from what can be considered ``promising,'' indicating that even the most reliable methods struggle to accurately detect domain shifts.

In contrast, the \texttt{ADD} results are mostly within an acceptable range across the different detectors. This suggests that while these algorithms may respond to changes relatively quickly, they often fail to accurately distinguish between true and false positives, leading to poor F1 performance.

\begin{figure}[!ht]
    \centering
    \begin{subfigure}{\linewidth}
    \centering
    
        \includegraphics[width=\linewidth]{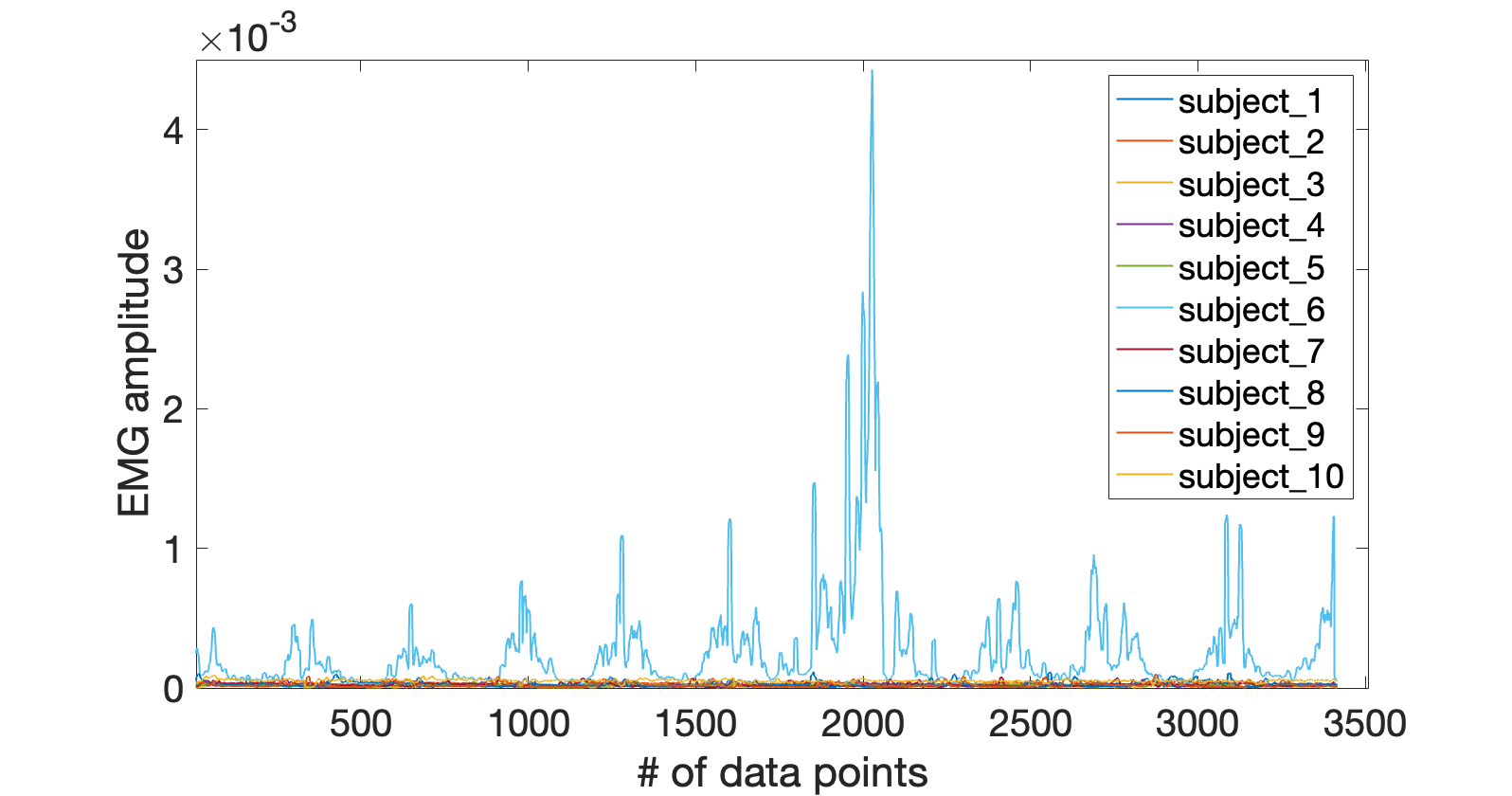}
        \label{fig:emg5}
        \caption{EMG\_2}
    \end{subfigure}
    \begin{subfigure}{\linewidth}
    \centering
    
        \includegraphics[width=\linewidth]{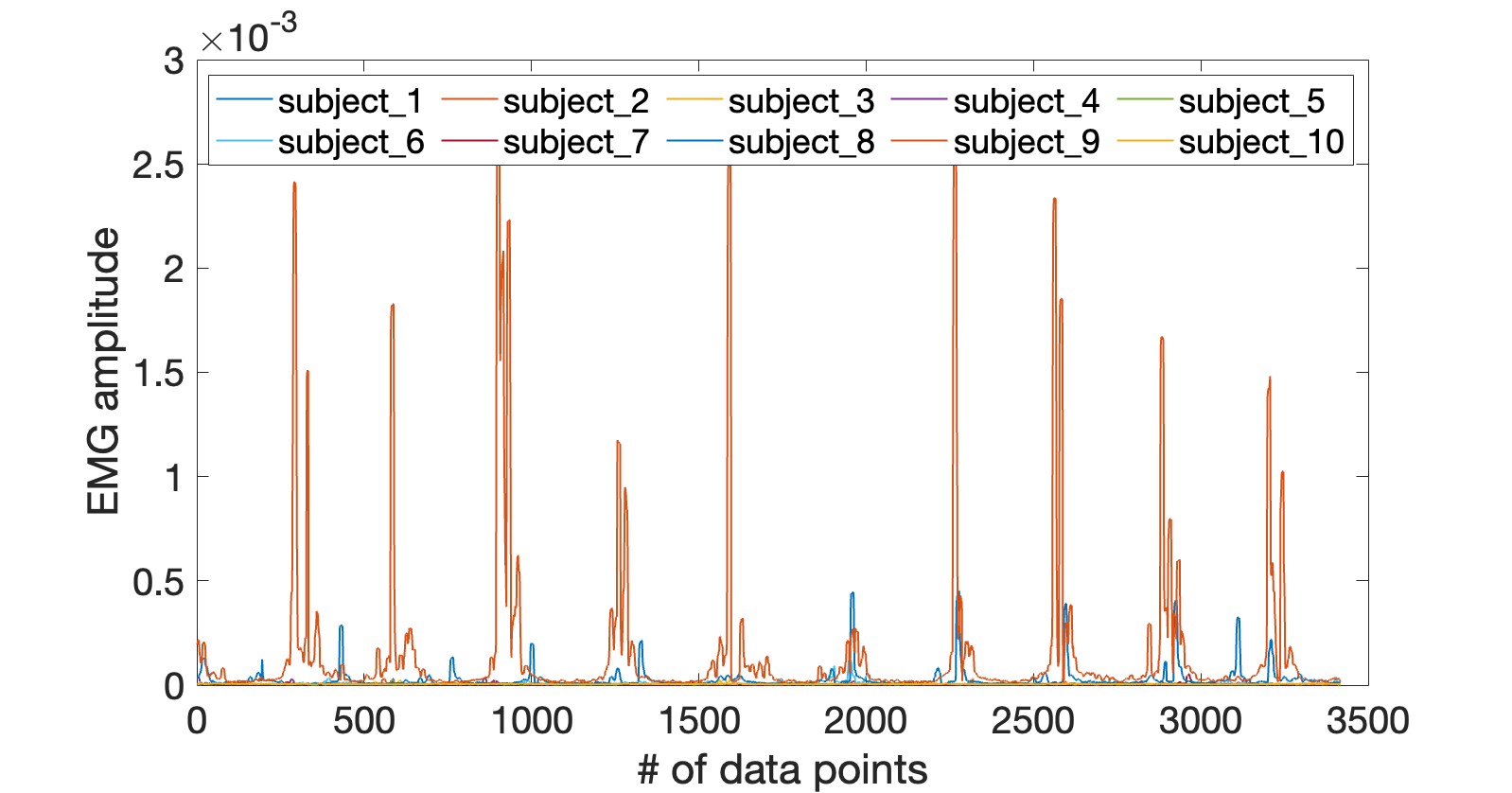}
        \label{fig:emg5}
        \caption{EMG\_5}
    \end{subfigure}%
    \caption{EMG Signals from 10 Subjects for Grasp 11 at Day 1, Time 1. Sub-figure a shows the amplitude for EMG\_2, and sub-figure b shows the amplitude for EMG\_5.}
    \label{fig:emg2_emg5}
\end{figure}

In summary, current drift detection methods appear insufficient for reliably identifying domain shifts in this type of data, and significant improvements are needed to develop methods that can better represent the time-varying data in the streaming setting and achieve both high accuracy and timely detection in real-world scenarios.
The performance presented in Table~\ref{tab:result} calls for a deeper understanding of the data. To highlight this, we present Fig.~\ref{fig:emg2_emg5} for further insights.

Fig.~\ref{fig:emg2_emg5}a and Fig.~\ref{fig:emg2_emg5}b present the EMG\_2 and EMG\_5 signals from all subjects recorded simultaneously. These figures illustrate how different subjects activate distinct muscle groups while performing the same actions.

In Fig.~\ref{fig:emg2_emg5}a, Subject 6 exhibits significantly stronger signals at the EMG\_2 than other subjects when repeating grasp 11. Conversely, Subject 9 shows greater muscle activation at the EMG\_5 while performing the same gesture, as depicted in Fig.~\ref{fig:emg2_emg5}b.

\begin{figure}[h]
    \centering
    \includegraphics[width=\linewidth]{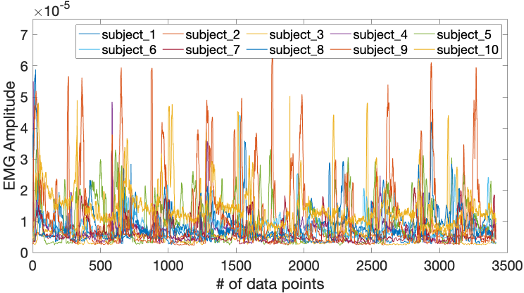}
    \caption{EMG\_7 Signals from 10 Subjects for Grasp 11 for Period 1}
    \label{fig:emg7}
\end{figure}

This observation suggests a preliminary conclusion: different individuals engage different muscle sections even when executing identical movements. This variation leads to the formation of distinct subject groups in Fig.~\ref{fig:different_subjects} and enhances the feasibility of detecting inter-subject differences.

Nonetheless, observing EMG\_7 gives us a better understanding why we get high false positive rate (see Fig.~\ref{fig:emg7}).
Although Fig.~\ref{fig:emg2_emg5} also highlights this issue, Fig.~\ref{fig:emg7} makes it even more evident -- signal fluctuations within a single subject's recording period are more pronounced than the variations observed across different subjects.

\section{Discussion and conclusion}\label{sec:Conclusion}

Due to the assumption that data stream learning consists of data with infinite length, most of the drift detection methods designed for data streams are based on statistical measurements and tests in order to ensure the efficiency of the detectors. 
This characteristic, however, results in a preference for capturing a large number of drifts in the data, i.e., the prominent changes within individual subjects.

A possible reason for this comes from the fact that DS are typically defined as infinite sequences of independently and identically distributed (i.i.d.) data items, whereas time series exhibit temporal dependencies between consecutive observations. 
While detecting subtle differences caused by domain shifts using DS techniques is intuitive, achieving a proper representation of the data by extracting features from the EMG/RMS data remains a significant challenge. 

The high efficiency and low memory requirements of streaming algorithms often conflict with the computational demands of non-linear processing techniques, such as Kernel PCA as discussed in Section~\ref{sec:Method}. Notably, experiments streaming variants of Kernel PCA -- such as Incremental PCA~\cite{ref_ipca} and Fair Streaming PCA~\cite{ref_spca} -- were also conducted, but no noticeable improvements were observed.

This work highlights several promising opportunities for future research, which can focus on the following directions:
\begin{enumerate}
    \item Incremental versions of non-linear decomposition models like Kernel PCA, enabling real-time adaptation without heavy computational costs;
    \item  Specialized drift detectors tailored for EMG signals which can significantly reduce false positives and improve robustness by accounting for the unique challenges of non-stationary EMG data;
    \item  Combining EMG with complementary data, such as accelerometer or grasp forces, that can enhance domain shift detection while improving overall system accuracy~\cite{gijsberts2014movement};
    \item  Incremental wavelet-based techniques can highlight inter-domain differences while smoothing in-domain variations, making detection of domain shifts easier; and 
        \item Developing methods for real-time recalibration of decoding models using detected domain-shifted data, thereby continuously adapting to evolving signal distributions and maintaining stable decoding performance in practical applications.
\end{enumerate}

\bibliographystyle{unsrt}
\bibliography{REF}
\end{document}